\documentclass[conference]{IEEEtran}
\IEEEoverridecommandlockouts
\usepackage{cite}
\usepackage{amsmath,amssymb,amsfonts}
\usepackage{algorithmic}
\usepackage{graphicx}
\usepackage{textcomp}
\usepackage{xcolor}
\usepackage{multirow}
\usepackage{bm}
\usepackage{makecell}
\def\BibTeX{{\rm B\kern-.05em{\sc i\kern-.025em b}\kern-.08em
    T\kern-.1667em\lower.7ex\hbox{E}\kern-.125emX}}
\begin{document}

\title{U3E: Unsupervised and Erasure-based Evidence Extraction for Machine Reading Comprehension}

\author{\IEEEauthorblockN{1\textsuperscript{st}Suzhe He}
\IEEEauthorblockA{\textit{Beijing Institute of Technology} \\
Beijing, China \\
suzhehe@bit.edu.cn}
\and
\IEEEauthorblockN{2\textsuperscript{nd}Shumin Shi*}
\IEEEauthorblockA{\textit{Beijing Institute of Technology} \\
Beijing, China  \\
bjssm@bit.edu.cn}
\and
\IEEEauthorblockN{3\textsuperscript{nd}ChengHao Wu}
\IEEEauthorblockA{
Beijing, China  \\
874635423@qq.com}
}

\maketitle

\begin{abstract}
More tasks in Machine Reading Comprehension (MRC) require, in addition to answer prediction, the extraction of evidence sentences that support the answer. However, the annotation of supporting evidence sentences is usually time-consuming and labor-intensive. In this paper, to address this issue and considering that most of the existing extraction methods are semi-supervised,  we propose an unsupervised evidence extraction method (U3E). U3E takes the changes after sentence-level feature erasure in the document as input, simulating the decline in problem-solving ability caused by human memory decline. In order to make selections on the basis of fully understanding the semantics of the original text, we also propose metrics to quickly select the optimal memory model for this input changes. To compare U3E with typical evidence extraction methods and investigate its effectiveness in evidence extraction, we conduct experiments on different datasets. Experimental results show that U3E is simple but effective, not only extracting evidence more accurately, but also significantly improving model performance.
\end{abstract}


\section{Introduction}
MRC has attracted more and more attention, and it can be roughly divided into two categories: extractive and non-extractive. Extractive MRC requires one or more paragraphs of text to be selected as the answer, such as SQuAD\cite{rajpurkar2016squad} and DROP\cite{dua2019drop}. Non-extractive MRC need reasoning ability more than the former. It includes multiple-choice MRC\cite{wang2020reco,ran2019option,zhu2021duma}, unanswerable questions\cite{rajpurkar2018know,zhang2020retrospective}, verification MRC\cite{clark2019boolq,zhang2021extract}. As shown in table \ref{tab1}, there is no doubt that evidence sentences can not only help the MRC model to predict the correct answer, but also help improve the interpretability of the model.
\cite{xie2022eider} effectively fuse the extracted evidence in reasoning to enhance the power of relation extraction. \cite{zhang2021summ} aIteratively performing long document segmentation enhances text generation performance.

However, it is difficult to label evidence sentences on large-scale datasets, especially for non-extractive MRC. Because a large number of these questions are not just extractive (e.g. 87\%  of  questions in RACE(\cite{lai2017race}, \cite{sun2019dream})). Answering Such questions may require more advanced reading comprehension skills such as single-sentence or multi-sentence reasoning skills (\cite{tu2020select}, \cite{chen2020multi}).

Considering the high cost of human-labeling evidence sentences, some recent efforts have been devoted to improving MRC by exploiting noisy evidence labels when training evidence extractors. Some work(\cite{lin2018denoising,min2018efficient}) generates remote tags using handcrafted rules and external sources, \cite{wang2019evidence} apply remote supervision to generate labels, denoising using a deep probabilistic logical learning framework. Some studies (\cite{choi2017coarse}) employ reinforcement learning (RL) to determine the labels of evidence, but this RL approach suffers from training instability. More distant monitoring techniques are also used to refine noisy labels. However, improving the evidence extractor is still challenging when golden evidence labels are not available.

So some recent works focus on weakly supervised extraction of evidence sentences, \cite{pruthi2020weakly} uses a small number of evidence annotations combined with a large number of document-level labels to select evidence, and\cite{niu2020self} uses a self-training approach that uses automatically generated evidence labels to supervise evidence selection in an iterative process. 

\begin{table}
\renewcommand\arraystretch{1.5}
\centering
\caption{Examples of Verified Reading Comprehension. Bold parts are evidence sentences. Answer 0 means the document does not support the option, 1 means it does.}\label{tab1}
\begin{tabular}{|l|}
\hline
{\bfseries O:} It is legal to own an ar15 in california. \\
{\bfseries D:} ... State of California. {\bfseries California's 2000 Assault Weapons ban} \\
{\bfseries  went further and banned AR-15s made by  ...}\\
{\bfseries A:} 0  \\

\hline
{\bfseries O:} Ferguson jenkins is in the hall of fame. \\
{\bfseries D:}   ... following his major league career. {\bfseries In 1991, Jenkins became } \\ 
{\bfseries the first Canadian to be inducted into the National Baseball} \\ 
{\bfseries  Hall of Fame.} \\
{\bfseries A:} 1  \\
\hline
\end{tabular}
\end{table}

In this paper, we propose an unsupervised method U3E for selecting evidence that is more in line with human intuition. Inspired by\cite{ju2021enhancing}, they change the characteristics of paragraph input (called erasure) by penalizing illogical predictions to improve the model. 

Instead of the loss value used in their work, we choose to use the change in the predicted value to reflect this memory-decay-like behavior because the predicted value is more sensitive. We also innovatively apply it to sentence level and the task of evidence extraction in MRC. Besides, in order to better adapt to this task, we propose an optimal memory model selection method BMC (balance model and changes). Experiments show that our method achieves good results.

\section{Related Work}
\subsection{Attribution Interpretation Methods}
In the field of hindsight, there are "variable importance" methods and gradient-based methods. The "variable importance" method\cite{li2016understanding,feng2018pathologies} refers to the difference in the prediction performance of a model when the value of a variable changes. In gradient-based methods, the magnitude of the gradient is used as the feature importance score. Gradient-based methods are suitable for differentiable models
\cite{sundararajan2017axiomatic}. Erasure\cite{ju2021enhancing} as a "variable importance" method, it is model independent. The advantage of the erasure method is that it is conceptually simple and can be optimized for well-defined objectives\cite{de2020decisions}.

\subsection{Evidence Extraction}
In the early MRC, some works focus on better representation of the features of the question and the context\cite{hu2017reinforced}, and constantly explore better fusion matching between them\cite{wang2018multi,ran2019option}. With the emergence of pre-training models (such as: BERT\cite{devlin2018bert}), some works want to understand the basis for the model to predict the answer. Extracting evidences in MRC is attracting increasing attention, although still quite challenging.

Evidence extraction aims to find evidence and relevant information for downstream processes in the task, which arguably improves the interpretability of the task. Evidence extraction is useful which is intuitive and becomes an important part of fact verification (\cite{ma2019sentence}, \cite{hanselowski2018ukp}), multiple-choice reading comprehension (\cite{yu2019inferential}), open-domain question answering (\cite{lin2018denoising}), multi-hop reading comprehension (\cite{inoue2021summarize}) , natural language inference (\cite{wang2017bilateral}), and a wide range of other tasks (\cite{chen2018fast}).

which can be roughly divided into two categories: one is supervised learning, which requires a lot of resources to manually label all the evidence sentence labels, such as: HOTPOTQA\cite{yang2018hotpotqa} Select evidence sentences on the basis of asking to answer their specific tasks, and work on it\cite{nishida2019answering} iteratively sorts the importance of sentences to select evidence sentences, \cite{min2019multi} decomposes the question and becomes a single-hop MRC extracts the evidence sentence while selecting the answer.

The second is semi-supervised learning. Because it is difficult to extract evidence sentences in non-extractive MRC, some works use semi-supervised methods to extract evidence, \cite{wang2019evidence} use remote supervision to generate imperfect labels, and then use deep probabilistic logic learning to remove noise. \cite{pruthi2020weakly} label and improve model performance by combining specific tasks with weakly supervised evidence extraction. Finally, on the basis of weakly supervised learning, \cite{wang2018r} use reinforcement learning to obtain better evidence extraction strategies.

Our method U3E tends to use an unsupervised method, complete the extraction task in stages. Adjust the erasure method so that the importance of sentences can be obtained explicitly. Compared with other methods, not only the cost is small, but also the effect is remarkable.

\section{Method}

\begin{figure}
\centering
\centerline{\includegraphics[width=1\linewidth]{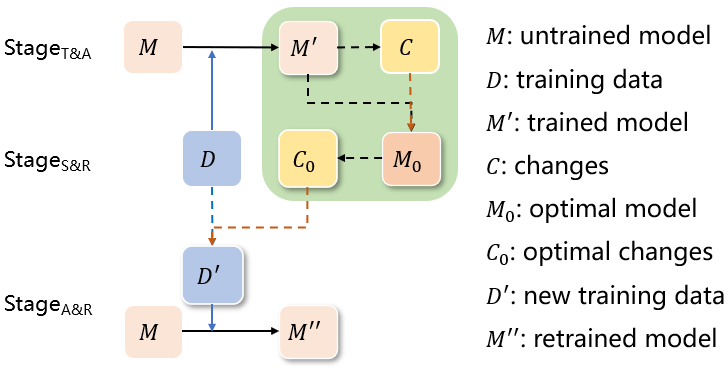}}
\caption{The overall structure of U3E, which includes three stages:
T\&A, S\&R, and A\&R. } \label{fig1}
\end{figure}

The overall architecture of U3E is shown in figure \ref{fig1}, which consists of three stages: \\
{\bfseries Train and Acquire (T\&A)}: train models according to the specific task and achieve changes. \\
{\bfseries Select and Reacquire (S\&R)}: select the optimal memory model according to our proposed BMC method and use the model to reacquire changes. \\
{\bfseries Apply and Retrain (A\&R)}: extract evidence through changes and retrain according to the evidence.\\
In the following specific implementation, we will explain these stages in order.

\subsection{Task Definition}

Assuming that each sample of the dataset can be formalized as follows: Given a reference document consisting of multiple sentences $ D = \left\{ S_1, S_2,..., S_m \right\} $ and a statement $ O $(If there is a question, then $O$ is represented as the concatenation of the question and the candidate) . The model should determine whether to support this statement according to the document, the support is marked as 1, otherwise it is marked as 0. It can also use to extract the evidence sentence set $ E=\left\{S_j,S_{j+1},...,S_{j+k-1} \right\} $, which contains $ k (< m) $ sentences in $ D $. 

\subsection{Train and Acquire}

\subsubsection{Task-specific Training}\label{tt}

We first train according to the specific task (here is the classified task), and then save the model $ M=\{M^1,M^2,...,M^x\} $ under all epochs, where x represents the largest epoch trained. The model structure during training is pretrained model\footnote{different pretrained models used on different datasets} and linear layer. The input is in " $ [CLS] + Option + [SEP] + Document + [SEP] $ " format. The hidden representation of the $ [CLS] $ token goes through a linear layer for binary classification to predict whether the document $D$ supports the sentence $O$:

\begin{figure*}
\centering
\centerline{\includegraphics[width=0.6\linewidth]{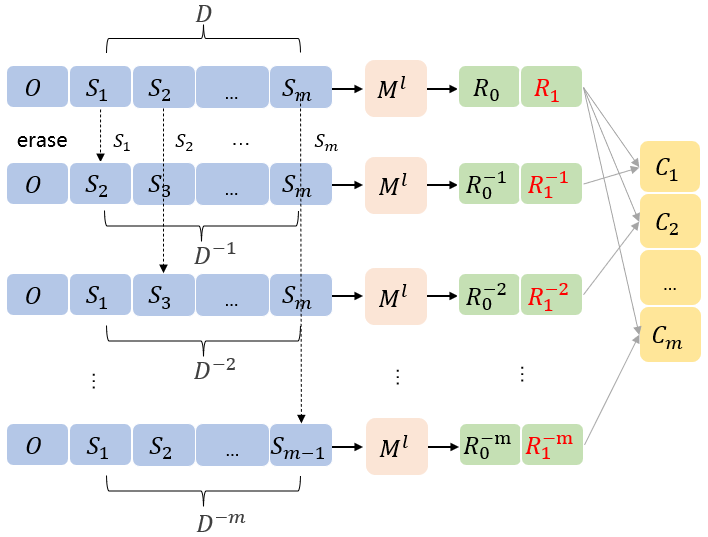}}
\caption{Implementation details for achieving changes. } \label{fig2}
\end{figure*}

\begin{equation}
\hat{y}  =  softmax(W_p h_{cls})
\end{equation}
where $ W_p $ is the model parameter and $ h_{cls} $ is the hidden representation of $ [CLS] $ token . The loss function is the cross entropy loss:

\begin{equation}
L = -\sum_{i=1}^{n}(y_i \cdot log(\hat{y}_i) + (1-y_i) \cdot log(1 - \hat{y}_i))
\quad j \in [1,N]
\end{equation}
where $ y_i $ represents the label of the sample $ (O_i, D_i) $, $ \hat{y} $ represents predicted value of the model, $N$ is the total number of samples.

\subsubsection{Change Acquisition}\label{ac}

In this part, we use the {\bfseries leave-one-out} method\cite{li2016understanding} to simulate memory decay and obtain the importance of each sentence. As shown in figure $\ref{fig2}$, $M^l$ is the model after the $l$th epoch training is completed, $ R $ represents the binary classification\footnote{without softmax} result predicted by model $ M^l $, where $ R_0 $ means the probability value of that document $ D $ does not support statement $ O $, otherwise, $ R_1 $ means support, and $ R_1 $ marked in red means that the correct answer for this example is support.

$ D^{-j} $ represents the new document obtained after erasing the j-th sentence, and the same $ R^{-j} $ represents the result obtained after replacing $ D^{-j} $  with $ D $ and then input the model $ M^l $ in the same format. The importance of the sentence $ C = \left\{c_1,c_2,..., c_m \right\}$ is calculated as follows:

\begin{equation}
c_j = abs( R_y - R_y^{-j} ) \quad j \in [1,m]
\end{equation}

Where $ y $ represents the sample label, $ abs $ represents the absolute value function and $ c_j $ indicates the importance of sentence $ j $.

We chose this method because the large positive effect on the result after erasure also means that it is most relevant to the statement $ O $. In other words, it is also a evidence, but it is an interference item, which can help the model improve its reasoning ability. More importantly, the predicted value changes can be more sensitive to the input changes.

\subsection{Select and Reacquire}

\subsubsection{Optimal Model Selection}\label{oms}

We have experimented with two optimal model selection methods as follows:
\paragraph{MTEST}

This method considers the generalization of the model, and specifically selects the model according to the maximum accuracy on the test set.

\paragraph{BMC}

According to research by \cite{jia2017adversarial}, neural reading comprehension models often exploiting spurious correlations in the data to predict answers. In order to prevent the model from biasing evidence sentences selection due to excessive pursuit of specific tasks, we propose an optimal model $M_0$ selection expression BMC (balance model and changes). 

The core of BMC method is to select the model that is most sensitive to the evidence sentence. So the BMC expression is as follows:

\begin{equation}
M_0 = \mathop{argmax}\limits_{M^l} (-\lambda \cdot Acc(M^l) + SC^l) \quad M^l \in M
\end{equation}

\begin{equation}
SC^l = \frac{1}{N} \cdot \sum_{i}^{N}{\frac{\sum C_{\sim k}(M^l)}{\sum C(M^l)}} 
\end{equation}

Where $M^l$ is the model after the l-th epoch is trained, $Acc$ represents the accuracy on the testing set and $\lambda$ is a hyperparameter. $k$ is the number of evidence sentences to be selected. 

\begin{table*}
\renewcommand\arraystretch{1.2}
\centering
\caption{Results in VGaokao.}\label{tab3}
\begin{center}
\begin{tabular}{|c|c|p{1cm}<{\centering}|}
\hline
 & {\bfseries Respective Acc.} & {\bfseries Acc.}\\
\hline
Full context &  $ \bm{-} $ &  64.46\\
\hline
Top 2 sentences by word vector &   $ \bm{-} $ &  63.79\\
Top 2 sentences by beam search &   $ \bm{-} $ &  63.63\\
\hline
Top 2 sentences by U3E$ _{mtest} $ &  69.91 / 64.46 & 63.93\\
Top 2 sentences by U3E$ _{BMC} $ & 99.64 / 64.18  &  64.54\\
Top 2 sentences by U3E$ _{MAX} $ & 99.51 / 62.32  & {\bfseries 65.96}\\
\hline
\end{tabular}
\end{center}
\end{table*}

$\sum C_{\sim k}(M^l)$ is the sum of the largest $k$ values in $C$, $\sum C(M^l)$ is the sum of all the values in $C$, the division of both represents the sensitivity of the model to the evidence sentence, we call it salient changes (SC).

Note that our model selection can be done simultaneously with training, so when training is over, the optimal model has already been selected.

\subsubsection{Changes Reacquisition}

The steps to reacquire the changes are similar to \ref{ac}, except that the model $M^l$ is replaced with the optimal model $M_0$. The final optimal changes are $C_0$.

\subsection{Apply and Retrain}\label{eer}

We choose the $ k $ sentences with the largest $ c_j $ in $ C_0 $ as the evidence sentence set $ E $ of statement $ O $. Then under the same model structure, the documents are replaced with ordered evidence sentences\footnote{consistent with the relative order of the original sentences} for retraining. 

This part is used to study whether the extracted evidence can improve the performance of the model(\ref{pim}), and is not needed in the evidence validity part(\ref{eva}).

\section{Experiments and Analysis}
\subsection{Dataset}

\textbf{VGaokao}\cite{zhang2021extract} is a validation dataset, which comes from the Chinese native speaker's Gaokao Chinese test and is a public dataset. It includes documents and statements, each statement requires at most two evidence sentences, 
and the answer yes/no directly indicates whether the document supports the statement.

\textbf{C$^{3}$}\cite{sun2019investigating} is a multiple-choice reading comprehension dataset. Here we use the C$^{3}$ dataset\footnote{https://ymcui.com/expmrc/} proposed in \cite{cui2022expmrc}, which is a competition dataset. It is necessary to construct pseudo-evidence labels for the training set, while the validation set has manually annotated evidence labels.

We use VGaokao to study the effect of generalization and the improvement of model performance, and C$^{3}$ to illustrate the accuracy of extracting evidence, since it has annotated test data.

\subsection{Baseline}

\subsubsection{Model structure}
 Both use the method of "pre-training model + linear layer", the difference is that the latter has two additional linear layers for evidence span extraction, they are used to predict where evidence begins and ends. The loss functions are all cross entropy loss. The former all-process pre-trained model uses chinese-roberta-wwm-ext, and the latter uses chinese-macbert-base.\footnote{https://huggingface.co/}
 Other differences in the C$^{3}$ are introduced in \ref{c3imp}.
 
 \subsubsection{Implementation}
 For dataset \textbf{VGaokao}, in order to obtain a more accurate experimental conclusion, during training, we select multiple most relevant sentences for each statement according to the static word vector\footnote{https://github.com/Embedding/Chinese-Word-Vectors \label{wv}}, ensuring that the length after splicing does not exceed the length limit of the pre-training model(average sentence length after operation is 7.793). Then the new document is formed after sorting according to the original position. During testing, we forecast the original document in blocks(step 128). Later, the subsequent experiments were conducted on the new document, and the tests were the same.

We use the following baselines.\\
-Word Vector: select the most relevant $k$ sentences from document $ D $ as evidence sentences for retraining based on word vectors\textsuperscript{\ref{wv}}. Here we use average pooling as sentence representation. \\
-Beam Search: beam search using the Hard Masking method proposed by \cite{zhang2021extract}, they delete the previously selected sentence information in the query, and iteratively select the evidence sentence. \\
-Full Context Train: in order to study the improvement of model performance, we use the original document $ D $(new document after filtering) for training.\\ 

 For dataset \textbf{C$^{3}$}, in order to exclude interference, we delete the data that is too long in the training set provided by \cite{cui-etal-2022-expmrc}. The model structure used is consistent with the baseline proposed in \cite{cui2022expmrc}.
Since the test set is not public, we evaluate it on the validation set. Here, only one sentence is required to be extracted as a proof sentence.

\subsection{Performance improvement} \label{pim}

\begin{table*}
\renewcommand\arraystretch{1.2}
\centering
\caption{Results in C$^{3}$.}\label{tab4}
\begin{tabular}{|c|c|c|c|c|}
\hline
Type of PrM & Method & {\bfseries ANS\_F1} & {\bfseries EVI\_F1} & {\bfseries ALL\_F1}\\
\hline
\multirow{2}{*}[-6pt]{Base} & RAW & 72.245 & $ \bm{-} $ & $ \bm{-} $ \\
 & WV  & 71.881 & 65.533 & 49.634 \\
 & {\bfseries U3E$ _{mtest} $} & 72.475 & {\bfseries 68.840}  & 53.385 \\
\hline
\multirow{2}{*}[-6pt]{Large} & RAW & 77.425 & $ \bm{-} $ & $ \bm{-} $ \\
 & WV  & 76.832 & 68.049 & 55.355 \\
 & {\bfseries U3E$ _{mtest} $} & 78.812 & {\bfseries 76.425}  & 62.720 \\
\hline
\end{tabular}
\end{table*}

\subsubsection{Implementation} \label{imp}

The results are shown in table \ref{tab3}. Except in Full data, {\bfseries Acc.} represents the test set accuracy on the original data. In other experiments, {\bfseries  Acc.} represents the accuracy on the test set during retraining after evidence extraction\ref{eer}. {\bfseries Acc.} for all experiments is obtained under the same experimental settings. The {\bfseries Respective Acc.} represents the accuracy of different methods of selecting the optimal model on the 
two dataset.

The model with the highest accuracy on the test set is selected for evidence extraction and retraining, and the result is subscripted by 'mtest'. For BMC, the hyperparameter $\lambda$ is set to 0.1, and the result is subscripted by 'BMC'. 

Then we also recorded the optimal result in 10 epoch and the test set accuracy at this time, which is subscripted by 'MAX'. 

\subsubsection{Result Analysis} \label{rea}

In general, the accuracy rate of using the U3E method after retraining is stronger than that of the general methods, indicating that U3E does have a better effect on the task of extracting key sentences. And U3E$ _{BMC} $ and U3E$ _{MAX} $ are higher than Full context, especially U3E$ _{MAX} $ is more than 1 point higher, which also means that our U3E method helps to improve the performance of the model.

U3E$ _{BMC} $ higher than U3E$ _{mtest} $ by 0.61 indicates that the BMC optimal model selection method is more effective than simply using the accuracy of the test set. This also suggests that excessive pursuit of task-specific accuracy may reduce the probability that the model predicts the answer from the evidence sentence. This is consistent with the study of \cite{jia2017adversarial}. However, U3E$ _{BMC} $ is still more than 1 point higher than the best effect U3E$ _{MAX} $, indicating that the BMC method still has room for improvement.

We observe that both U3E$ _{BMC}$ and U3E$ _{MAX} $ have more than 90 accuracy on the training set, indicating that the U3E method needs to have sufficient knowledge of the training set.

\subsection{Evidence Validity} \label{eva}

To further demonstrate the accuracy of our method for extracting evidence, we conduct experiments on dataset C$^{3}$.

\subsubsection{Implementation}\label{c3imp}

Considering that the candidate answers of the c dataset are not long, we do not need to perform evidence extraction for each candidate, so we do the splicing of "document, question and candidate" for each candidate, and then enter the model together.

Considering that there are two tasks, we design the total loss function as follows:
 
\begin{equation}
Loss_{total} = Loss_{ans} + \alpha \cdot Loss_{evi} \quad \alpha \in [0,1]
\end{equation}

Where $ Loss_{ans} $ is the loss of multiple answers, and $ Loss_{evi} $ is the loss of evidence span. We record the maximum experimental results every 0.1 from 0 to 1.

 In this part, we use U3E's MTEST method for evidence extraction, which is to extract evidence according to the best model on the validation set. Note that here we only extract evidence for training set.
 
 As shown in Table \ref{tab4}, RAW represents the answer prediction accuracy without evidence extraction task. WV means using static word vectors to extract evidence, and U3E$ _{mtest} $ means using U3E method to extract evidence, both of which select evidence sentences according to the correct answers and questions. ANS\_F1 is the accuracy of multiple-choice answers, EVI\_F1 is the accuracy of evidence, and ALL\_F1 is the accuracy of the combination of the two. The evaluation method is provided by \cite{cui2022expmrc}.
 
\subsubsection{Result Analysis}

As shown in Table \ref{tab4}, the evidence obtained using our method greatly improves the upper limit of the evidence accuracy rate, from 65.533 to 68.840, and simultaneous training can help improve the accuracy of multiple-choice answers, from 72.245 to 72.475. Moreover, according to the above research, method NT$ _{test} $ does not use the optimal model, and a greater improvement is expected.

Moreover, when the pre-training model is larger\footnote{ https://huggingface.co/hfl/chinese-macbert-large}, we observe more significant changes. U3E not only improves the evidence prediction by more than 8 points, but also improves the multi choice answer prediction by more than 1 point compared with the original training data.

\begin{table*}
\renewcommand\arraystretch{1.5}
\centering
\caption{Evidence extraction results. WV is the static word vector,and \\ U3E is the U3E method. The following is the Chinese translation.}\label{tab5}
\begin{center}
\begin{tabular}{p{12cm}}
\hline
\textbf{ Query:} What might their relationship be?  \\
\textbf{ WV: } Woman: Yes, it's been more than ten years since we saw each other. \\
\textbf{U3E:} Man: But you haven't changed at all, just like when you were in \textbf{college}. \\
\hline
\textbf{ Query:} How do animals behave when they sense low-frequency sound waves?\\
\textbf{ WV: } This friction creates a low-frequency sound wave that is lower than what the human hearing can perceive.\\
\textbf{ U3E: } Those animals with very sensitive senses will \textbf{panic} when they feel this low sound wave, and even ...\\
\hline
\textbf{ Query:} What kind of teacher did Zhang Liyong become?\\
\textbf{ WV: } Zhang Liyong did not become a teaching assistant in the English Department of Tsinghua University, but many people asked him to be a teacher.\\
\textbf{ U3E: } He is an English \textbf{tutor} for a third-year student.\\
\hline
\end{tabular}
\end{center}
\end{table*}

We also compare the specific effects of using word vectors and U3E, as shown in Table \ref{tab5}, the former tends to select more similar sentences, which not only fail to answer the question, but also mislead the model. The latter makes full use of the favorable conditions of the model "understanding the semantics of the original text" and selects the sentence, which has the largest change in the confidence of the correct answer after the memory declines, as the evidence sentence. It's more intuitive and more accurate.

\section{Conclusion}

We propose an erasure-based method called U3E for unsupervised evidence extraction. U3E build a model based on the reduced ability to process problems due to memory loss in humans. In order to select the optimal memory model quickly, a calculation method called BMC is proposed. The experimental results show that U3E not only accurately extracts evidence, but also improves model performance.

Next, we are going to mine a more precise relationship between model accuracy and changes to quickly complete the selection of the optimal model, and apply the U3E to other NLP tasks such as natural language inference.

\bibliographystyle{unsrt}
\bibliography{myArticle}

\end{document}